\documentclass[10pt,twocolumn,letterpaper]{article}

\usepackage{cvpr}
\usepackage{times}
\usepackage{epsfig}
\usepackage{graphicx}
\usepackage{amsmath}
\usepackage{amssymb}
\usepackage{tikz}
\usepackage{float}
\usepackage{bbm}
\usepackage{subfig}
\usepackage{stfloats}

\usepackage[pagebackref=true,breaklinks=true,letterpaper=true,bookmarks=false]{hyperref}

\cvprfinalcopy

\ifcvprfinal\fi
\begin{document}

\title{Painting Outside the Box: Image Outpainting with GANs}

\author{Mark Sabini and Gili Rusak\\
Stanford University\\
{\tt\small \{msabini, gilir\}@cs.stanford.edu}}

\maketitle

\begin{abstract}
    The challenging task of image outpainting (extrapolation) has received comparatively little attention in relation to its cousin, image inpainting (completion). Accordingly, we present a deep learning approach based on \cite{iizuka2017} for adversarially training a network to hallucinate past image boundaries. We use a three-phase training schedule to stably train a DCGAN architecture on a subset of the Places365 dataset. In line with \cite{iizuka2017}, we also use local discriminators to enhance the quality of our output. Once trained, our model is able to outpaint $128\times 128$ color images relatively realistically, thus allowing for recursive outpainting. Our results show that deep learning approaches to image outpainting are both feasible and promising.
\end{abstract}

\section{Introduction}
The advent of adversarial training has led to a surge of new generative applications within computer vision. Given this, we aim to apply GANs to the task of \textbf{image outpainting} (also known as image extrapolation). In this task, we are given an $m \times n$ source image $I_s$, and we must generate an $m \times n + 2k$ image $I_o$ such that:
\begin{itemize}
    \item $I_s$ appears in the center of $I_o$
    \item $I_o$ looks realistic and natural
\end{itemize}

Image outpainting has been relatively unexplored in literature, but a similar task called \textbf{image inpainting} has been widely studied. In contrast to image outpainting, image inpainting aims to restore deleted portions in the interiors of images. Although image inpainting and outpainting appear to be closely related, it is not immediately obvious whether techniques for the former can be directly applied to the latter.

Image outpainting is a challenging task, as it requires extrapolation to unknown areas in the image with less neighboring information. In addition, the output images must appear realistic to the human eye. One common method for achieving this in image inpainting involves using GANs \cite{iizuka2017}, which we aim to repurpose for image outpainting. As GANs can be difficult to train, we may need to modify the archetypal training procedure to increase stability.

Despite the challenges involved in its implementation, image outpainting has many novel and exciting applications. For example, we can use image outpainting for panorama creation, vertically-filmed video expansion, and texture creation.

In this project, we focus on achieving image outpainting with $m = 128$, $n = 64$, and $k = 32$.

\section{Related Work}
One of the first papers to address image outpainting used a data-driven approach combined with a graph representation of the source image \cite{wang2014}. Although the researchers were able to achieve realistic results, we hope to apply adversarial training for potentially even better results.

A key implementation of image inpainting using deep learning by Pathak et al. introduced the notion of a Context Encoder, a CNN trained adversarially to reconstruct missing image regions based on surrounding pixels \cite{pathak2016context}. The results presented were relatively realistic, but still had room for visual improvement.

Iizuka et al. improved on the Context Encoder approach for image inpainting by adding a second discriminator that only took as input the inpainted patch and its immediate surroundings \cite{iizuka2017}. This ``local" discriminator, combined with the already-present ``global" discriminator, allowed for the researchers to achieve very visually convincing results. As a result, this approach is a promising starting point for achieving image outpainting.

Finally, recent work in image inpainting by Liu et al. \cite{liu2018image} utilized partial convolutions in conjunction with the perceptual and style loss first introduced by Gatys et al. \cite{gatys2015neural}. Using these techniques, the researchers were able to achieve highly realistic results with a fraction of the training required by \cite{iizuka2017}. In addition, the researchers provided various quantitative metrics that will be useful for evaluating our models' performance.

\section{Dataset}
As a sanity check for the outpainting model architecture, we expect our model to be able to overfit on a single $128 \times 128$ color image of a city. We use a $128 \times 128$ image as opposed to the $512 \times 512$ image size from \cite{iizuka2017} to speed up training. For this experiment, we use the same single image for training and testing.

\begin{figure}
    \centering
    \includegraphics[scale=0.5]{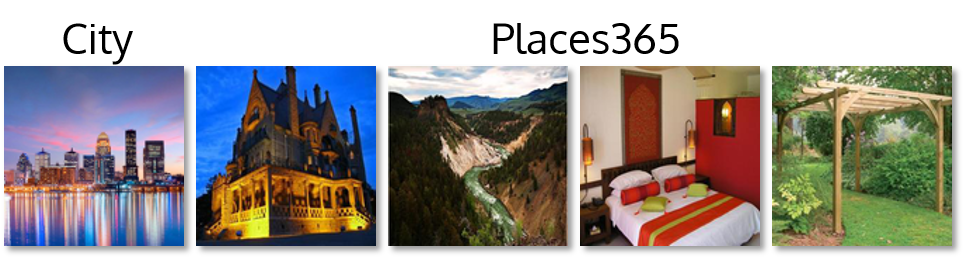}
    \caption{City image and sample images from Places365 dataset.}
    \label{fig:places365}
\end{figure}

Our primary dataset for image outpainting is composed of $36,500$ $256 \times 256$ images from the Places 365 dataset \cite{zhou2017places}. We downsampled these images to $128 \times 128$. This dataset is composed of a diverse set of landscapes, buildings, rooms, and other scenes from everyday life, as shown in Figure \ref{fig:places365}. For training, we held out $100$ images for validation, and trained on the remaining $36,400$ images.

\subsection{Preprocessing}
In order to prepare our images for training, we use the following preprocessing pipeline. Given a training image $I_{\mathrm{tr}}$, we first normalize the images to $I_{n} \in [0,1]^{128 \times 128 \times 3}$. We define a mask $M \in  \{0,1\}^{128 \times 128}$ such that $M_{ij} = 1 - \mathbf{1} [ 32 \leq j < 96]$ in order to mask out the center portion of the image. Next, we compute the mean pixel intensity $\mu$, over the unmasked region $I_{n} \odot (1-M)$. Afterwards, we set the outer pixels of each channel to the average value $\mu$. Formally, we define $I_{m} = \mu \cdot M + I_{n} \odot (1-M)$. In the last step of preprocessing, we concatenate $I_m$ with $M$ to produce $I_p \in [0, 1]^{128 \times 128 \times 4}$. Thus, as the result of preprocessing $I_{\mathrm{tr}}$, we output $(I_n, I_p)$.

\section{Methods}
\subsection{Training Pipeline}
We adopt a DCGAN architecture $(G, D)$ similar to that used by Iizuka et al. Here, the generator $G$ takes the form of an encoder-decoder CNN, while the discriminator $D$ uses strided convolutions to repeatedly downsample an image for binary classification \cite{iizuka2017}.

\begin{figure}
    \centering
    \includegraphics[scale=0.5]{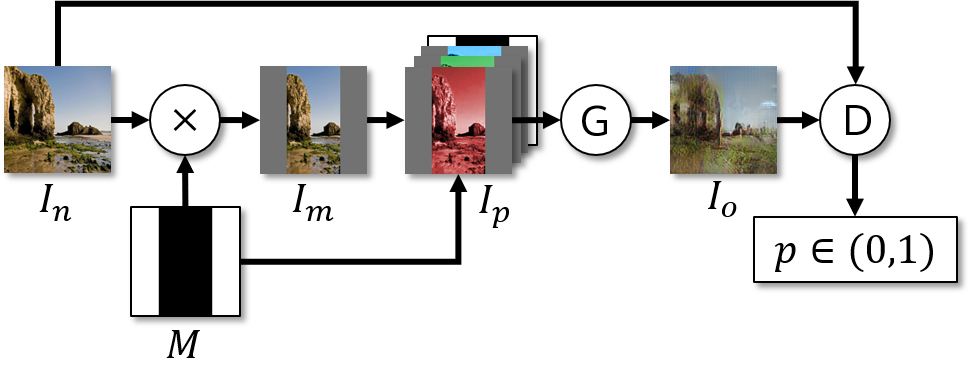}
    \caption{Training pipeline}
    \label{fig:pipeline}
\end{figure}

In each iteration of training, we randomly sample a minibatch of training data. As shown in Figure \ref{fig:pipeline}, for each training image $I_\mathrm{tr}$, we preprocess $I_\mathrm{tr}$ to get $I_n$ and $I_p$, as previously described. We run the generator on $I_p$ to get the outpainted image $I_o = G(I_p) \in [0, 1]^{128\times 128\times 3}$. Afterwards, we run the discriminator to classify the ground truth ($I_n$) and outpainted image ($I_o$). We compute losses and update parameters according to our training schedule, which will be discussed next.

\subsection{Training Schedule}
In order to facilitate and stabilize training, we utilize the three-phase training procedure presented by \cite{iizuka2017}. In this scheme, we define three loss functions:
\begin{equation}
    \mathcal{L}_{\mathrm{MSE}}(I_n, I_p) = \|M \odot (G(I_p) - I_n)\|^2_2
\end{equation}
\begin{equation}
    \mathcal{L}_D(I_n, I_p) = -\left[\log D(I_n) + \log(1 - D(G(I_p)))\right]
\end{equation}
\begin{equation}
    \mathcal{L}_G(I_n, I_p) = \mathcal{L}_{\mathrm{MSE}}(I_n, I_p) -\alpha \cdot \log D(G(I_p))
\end{equation}

The first phase of training (termed \textbf{Phase 1}) conditions the generator by updating the generator weights according to $\mathcal{L}_{\mathrm{MSE}}$ for $T_1$ iterations. The next phase (termed \textbf{Phase 2}) similarly conditions the discriminator by updating the discriminator weights according to $\mathcal{L}_D$ for $T_2$ iterations. The rest of training (termed \textbf{Phase 3}) proceeds for $T_3$ iterations, in which the discriminator and generator are trained adversarially according to $\mathcal{L}_D$ and $\mathcal{L}_G$, respectively. In $\mathcal{L}_G$, $\alpha$ is a tunable hyperparameter trading off the MSE loss with the standard generator GAN loss.

\subsection{Network Architecture}
Due to computational restrictions, we propose an architecture for outpainting on Places365 that is shallower but still conceptually similar to that by Iizuka et al. \cite{iizuka2017}. For the generator $G$, we still maintain the encoder-decoder structure from \cite{iizuka2017}, as well as dilated convolutions to increase the receptive field of neurons and improve realism.

For the discriminator $D$, we still utilize local discriminators as in \cite{iizuka2017}, albeit modified for image outpainting. Specifically, say the discriminator is run on an input image $I_d$ (equivalent to either $I_n$ or $I_o$ during training). In addition, define $I_\ell$ to be the left half of $I_d$, and $I_r'$ to be the right half of $I_d$, flipped along the vertical axis. This helps to ensure that the input to $D_\ell$ always has the outpainted region on the left. Then, to generate a prediction on $I_d$, the discriminator computes $D_g(I_d)$, $D_\ell(I_\ell)$, and $D_\ell(I_r')$. These three outputs are then fed into the concatenator $C$, which produces the final discriminator output $p = C(D_g(I_d) \parallel D_\ell(I_\ell) \parallel D_\ell(I_r'))$.

We describe the layers of our architecture in Figures \ref{fig:gen} and \ref{fig:disc}. Here, $f$ is the filter size, $\eta$ is the dilation rate, $s$ is the stride, and $n$ is the number of outputs. In all networks, every layer is followed by a ReLU activation, except for the final output layer of the generator and concatenator: these are followed by a sigmoid activation.

\begin{figure}
\caption{Generator, $G$}
\label{fig:gen}
\begin{center}
\begin{tabular}{ccccc}
\hline
Type & $f$ & $\eta$ & $s$ & $n$ \\ \hline
CONV & $5$ & $1$ & $1$ & $64$ \\ \hline
CONV & $3$ & $1$ & $2$ & $128$ \\
CONV & $3$ & $1$ & $1$ & $256$ \\ \hline
CONV & $3$ & $2$ & $1$ & $256$ \\
CONV & $3$ & $4$ & $1$ & $256$ \\
CONV & $3$ & $8$ & $1$ & $256$ \\
CONV & $3$ & $1$ & $1$ & $256$ \\ \hline
DECONV & $4$ & $1$ & $\frac{1}{2}$ & $128$ \\
CONV & $3$ & $1$ & $1$ & $64$ \\
OUT & $3$ & $1$ & $1$ & $3$ \\ \hline
\end{tabular}
\end{center}
\end{figure}

\begin{figure}
\caption{Discriminator, $D$}
\label{fig:disc}
\begin{center}
\subfloat[Global Discriminator, $D_g$]{
    \begin{tabular}{cccc}
    \hline
    Type & $f$ & $s$ & $n$ \\ \hline
    CONV & $5$ & $2$ & $32$ \\
    CONV & $5$ & $2$ & $64$ \\
    CONV & $5$ & $2$ & $64$ \\
    CONV & $5$ & $2$ & $64$ \\
    CONV & $5$ & $2$ & $64$ \\ \hline
    FC & - & - & $512$ \\ \hline
    \end{tabular}
}
\quad
\subfloat[Local Discriminator, $D_\ell$]{
    \begin{tabular}{cccc}
    \hline
    Type & $f$ & $s$ & $n$ \\ \hline
    CONV & $5$ & $2$ & $32$ \\
    CONV & $5$ & $2$ & $64$ \\
    CONV & $5$ & $2$ & $64$ \\
    CONV & $5$ & $2$ & $64$ \\ \hline
    FC & - & - & $512$ \\ \hline
    \end{tabular}
}
\newline
\subfloat[Concatenation layer, $C$]{
    \begin{tabular}{cccc}
    \hline
    Type & $f$ & $s$ & $n$ \\ \hline
    concat & - & - & $1536$ \\
    FC & - & - & $1$ \\ \hline
    \end{tabular}
}
\end{center}
\end{figure}

\subsection{Evaluation Metrics}
Although the output of the generator is best evaluated qualitatively, we still utilize RMSE as our primary quantitative metric. Given a ground truth image $I_{\mathrm{tr}} \in [0, 255]^{128\times 128\times 3}$ and a normalized generator output image $I_o' = 255 \cdot I_o \in [0, 255]^{128\times 128\times 3}$, we define the RMSE as:
$$\mathrm{RMSE}(I_{\mathrm{tr}}, I_o') = \sqrt{\frac{1}{|\mathrm{supp}(M)|}\sum\limits_{i, j, k}(M \odot (I_{\mathrm{tr}} - I_o'))^2_{ijk}}$$

\subsection{Postprocessing}
In order to improve the quality of the final outpainted image, we apply slight postprocessing to the generator's output $I_o$. Namely, after renormalizing $I_o$ via $I_o' = 255 \cdot I_o$, we blend the unmasked portion of $I_{\mathrm{tr}}$ with $I_o'$ using OpenCV's seamless cloning, and output the blended outpainted image $I_{\mathrm{op}}$.

\section{Results}
\subsection{Overfitting on a Single Image}
In order to test our architecture and training pipeline, we ran an initial baseline on the single city image. The network was able to overfit to the image, achieving a final RMSE of only $0.885$. This suggests that the architecture is sufficiently complex, and likely able to be used for general image outpainting.

\subsection{Outpainting on Places365}
We trained our architecture using only a global discriminator on Places365. As in \cite{iizuka2017}, we used a $18/2/80$ split for the different phases of training, and ran three-phase training with $T_1 = 40,950$, $T_2 = 4,550$, $T_3 = 182,000$, and $\alpha = 0.0004$. As we utilized a batch size of $16$, this schedule corresponded to approximately $100$ epochs and $40$ hours of training on a K80 GPU.\footnote{Animations of the generator's output during the course of training are available at \url{http://marksabini.com/cs230-outpainting/}.}

\begin{figure*}
    \centering
    \includegraphics[scale=0.5]{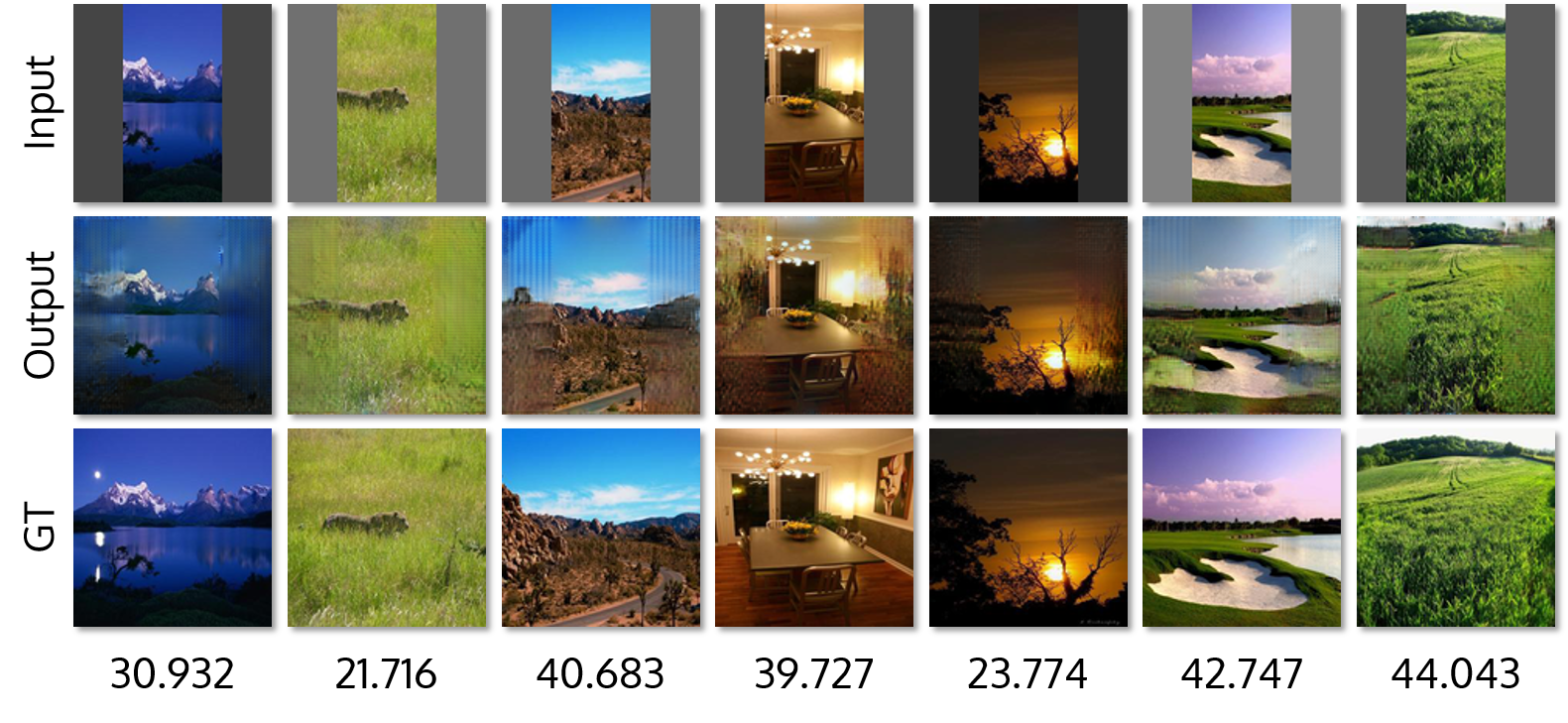}
    \caption{Outpainting results for a sample of held-out images in the validation set, shown alongside the original ground truth. These results were produced without the aid of local discriminators. The RMSE between each $(\textrm{ground truth}, \textrm{output})$ pair is displayed below each column.}
    \label{fig:results}
\end{figure*}

At the end of training, we fed images from the validation set through the outpainting pipeline. The final results, along with their RMSE values, are shown in Figure \ref{fig:results}. As seen in the third example of Figure \ref{fig:results}, the network does not merely copy visual features and lines during outpainting. Rather, it learns to hallucinate new features, as shown by the appearance of a house on the left hand side.

\begin{figure}
    \centering
    \includegraphics[scale=0.5]{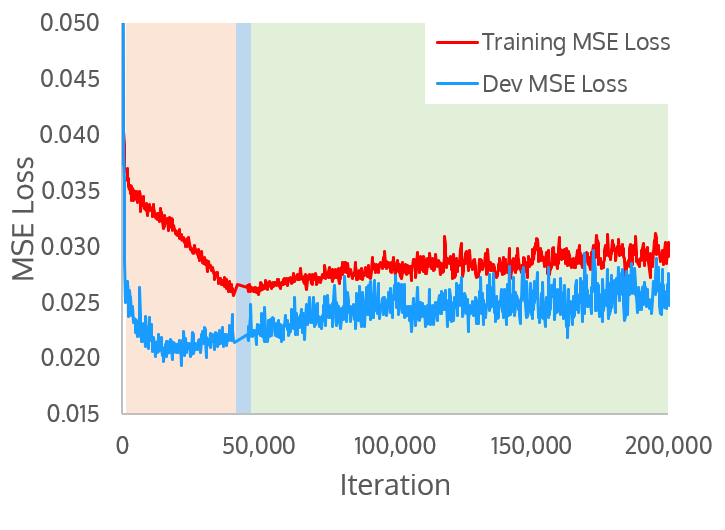}
    \caption{Training and dev MSE loss for training on Places365 with only a global discriminator. The orange, blue, and green sections represent Phase 1, 2, and 3 of training, respectively.}
    \label{fig:loss_plot}
\end{figure}

Figure \ref{fig:loss_plot} shows the training and dev MSE loss of this full run. In Phase 1, the MSE loss decreases quickly as it is directly optimized. On the other hand, in Phase 3, the MSE loss increases slightly as we optimize the joint loss (3).

\subsection{Local Discriminator}
In order to attempt to improve the quality of our results, we also trained our architecture using a local discriminator on Places365. Due to slower training, we trained this augmented architecture using three-phase training with $T_1 = 20,000$, $T_2 = 4,000$, $T_3 = 95,000$, and $\alpha = 0.0004$. With a batch size of $16$, this schedule corresponded to approximately $50$ epochs and $26$ hours of training on a K80 GPU.

\begin{figure}
    \centering
    \includegraphics[scale=0.5]{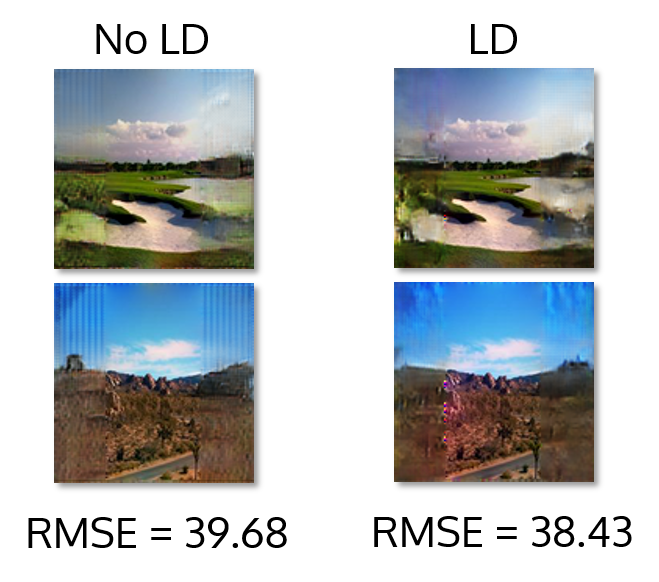}
    \caption{Training with local discriminators (LD) reduced vertical banding and improved color fidelity, but increased point artifacts and training time.}
    \label{fig:local_discriminator}
\end{figure}

After training, we compared the visual quality and the RMSE of images outpainted with and without the aid of a local discriminator. As seen in Figure \ref{fig:local_discriminator}, we observed that training with a local discriminator reduced vertical banding, improved color fidelity, and achieved a lower RMSE. This is likely because the local discriminator is able to focus on only the outpainted regions. However, the local discriminator caused training to proceed roughly $60\%$ slower, and tended to introduce more point artifacts.

\subsection{Significance of Dilated Convolutions}

\begin{figure}
    \centering
    \includegraphics[scale=0.5]{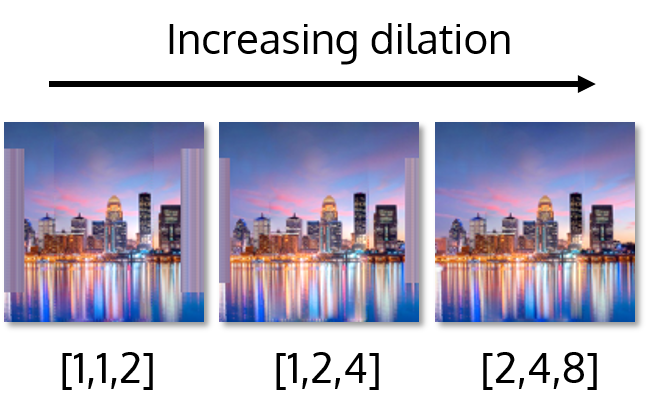}
    \caption{Effect of dilated convolutions. $[i, j, k]$ represent $\eta$ for layers $4$, $5$, and $6$ in the generator, respectively.}
    \label{fig:dilated_conv}
\end{figure}

We tuned the architecture by experimenting with different dilation rates for the dilated convolution layers of the generator. We attempted to overfit our architecture on the single city image with various layer hyperparameters. As shown in Figure \ref{fig:dilated_conv}, with insufficient dilation, the network fails to outpaint due to a limited receptive field of the neurons. With increased dilations, the network is able to reconstruct the original image.

\subsection{Recursive Outpainting}

\begin{figure}
    \centering
    \includegraphics[scale=0.5]{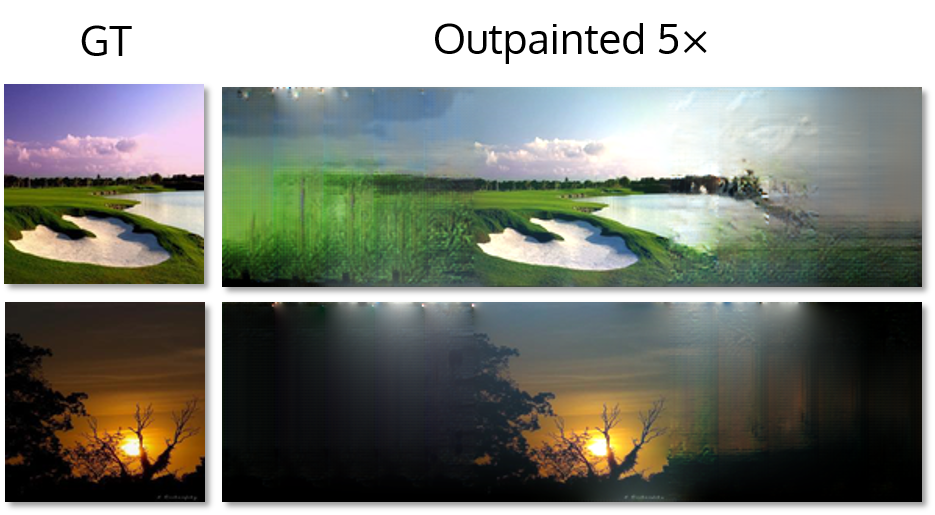}
    \caption{Each right image is the result of recursively-outpainting the corresponding left image five times.}
    \label{fig:recursive}
\end{figure}

An outpainted image $I_o$ can be fed again as input to the network after expanding and padding with the mean pixel value. In Figure \ref{fig:recursive}, we repeat this process recursively five times, expanding an image's width up to a factor of $3.5$. As expected, the noise tends to compound with successive iterations. Despite this, the model successfully learns the general textures of the image and extrapolates the sky and landscape relatively realistically. 

\section{Conclusions}
We were able to successfully realize image outpainting using a deep learning approach. Three-phase training proved to be crucial for stability during GAN training. In addition, dilated convolutions were necessary to provide sufficient receptive field to perform outpainting. The results from training with only a global discriminator were fairly realistic, but augmenting the network with a local discriminator generally improved quality. Finally, we investigated recursive outpainting as a means of arbitrarily extending an image. Although image noise compounded with successive iterations, the recursively-outpainted image remained relatively realistic.

\section{Future Work}
Going forward, there are numerous potential improvements for our image outpainting pipeline. To boost the performance of the model, the generator loss could be augmented with perceptual, style, and total variation losses \cite{gatys2015neural}. In addition, the architecture could be modified to utilize partial convolutions, as explored in \cite{liu2018image}. To further stabilize training, the Wasserstein GAN algorithm could be incorporated into three-phase training \cite{arjovsky2017wasserstein}. With the aid of sequence models, the image outpainting pipeline could even be conceivably extended to perform video outpainting.

\section{Acknowledgements}
We would like to thank Jay Whang for his continued mentorship throughout the course of this project.

We would also like to thank Stanford University's CS 230 (Deep Learning), taught by Professor Andrew Ng and Kian Katanforoosh, for offering a great learning opportunity.

In addition, we would like to thank Amazon Web Services for providing GPU credits.

\section{Supplementary Material}

The code and accompanying poster for our project can be found at \url{https://github.com/ShinyCode/image-outpainting}.

\nocite{*}
{\small
\bibliographystyle{ieee}
\bibliography{egbib}
}

\end{document}